\documentclass[journal,twoside,web]{IEEEtran}
\usepackage{generic}
\usepackage{cite}
\usepackage{amsmath,amssymb,amsfonts}
\usepackage{algorithmic}
\usepackage{graphicx}
\usepackage{textcomp}
\usepackage{xcolor}
\usepackage{color}
\usepackage{multirow}
\usepackage{mathtools}
\usepackage[ruled, lined, linesnumbered, commentsnumbered, longend]{algorithm2e}

\usepackage{tikz}
\usepackage{textcomp}
\usepackage{hyperref}
\usepackage{lipsum}

\newcommand\copyrighttext{%
  \footnotesize \textcopyright 2021 IEEE. Personal use of this material is permitted.
  Permission from IEEE must be obtained for all other uses, in any current or future
  media, including reprinting/republishing this material for advertising or promotional
  purposes, creating new collective works, for resale or redistribution to servers or
  lists, or reuse of any copyrighted component of this work in other works.
  DOI: \href{<https://ieeexplore.ieee.org/document/9326423>}{10.1109/JBHI.2021.3052320}}
\newcommand\copyrightnotice{%
\begin{tikzpicture}[remember picture,overlay]
\node[anchor=south] at (current page.south) {\fbox{\parbox{\dimexpr\textwidth-\fboxsep-\fboxrule\relax}{\copyrighttext}}};
\end{tikzpicture}%
}

\def\BibTeX{{\rm B\kern-.05em{\sc i\kern-.025em b}\kern-.08em
    T\kern-.1667em\lower.7ex\hbox{E}\kern-.125emX}}
\begin{document}
\title{DSAL: Deeply Supervised Active Learning from Strong and Weak Labelers for Biomedical Image Segmentation}
    \author{Ziyuan Zhao, Zeng Zeng, \IEEEmembership{Senior Member, IEEE}, Kaixin Xu, Cen Chen, Cuntai Guan, \IEEEmembership{Fellow, IEEE}
\thanks{Manuscript received September 30, 2020; revised November 16, 2020 and December 13, 2020; accepted January 11, 2021. Date of publication November 13, 2018; date of current version November 6, 2019. The  research  was  funded  by  the  Singapore-China NRF-NSFC Grant, No. NRF2016NRF-NSFC001-111. (Corresponding
author: Zeng Zeng, Cen Chen.)}
\thanks{Ziyuan Zhao, Zeng Zeng, Kaixin Xu, and Cen Chen are with Infocomm for Research Institute, Agency for
Science, Technology and Research, Singapore. (e-mail: \{zhaoz, zengz, xuk, chenc\}@i2r.a-star.edu.sg).
}
\thanks{Cuntai Guan is with Nanyang Technological University, Singapore. (e-mail: ctguan@ntu.edu.sg).}
}

\maketitle

\copyrightnotice

\begin{abstract}
Image segmentation is one of the most essential biomedical image processing problems for different imaging modalities, including microscopy and X-ray in the Internet-of-Medical-Things (IoMT) domain. However, annotating biomedical images is knowledge-driven, time-consuming, and labor-intensive, making it difficult to obtain abundant labels with limited costs. Active learning strategies come into ease the burden of human annotation, which queries only a subset of training data for annotation. Despite receiving attention, most of active learning methods still require huge computational costs and utilize unlabeled data inefficiently. They also tend to ignore the intermediate knowledge within networks. In this work, we propose a deep active semi-supervised learning framework, DSAL, combining active learning and semi-supervised learning strategies. In DSAL, a new criterion based on deep supervision mechanism is proposed to select informative samples with high uncertainties and low uncertainties for strong labelers and weak labelers respectively. The internal criterion leverages the disagreement of intermediate features within the deep learning network for active sample selection, which subsequently reduces the computational costs. We use the proposed criteria to select samples for strong and weak labelers to produce oracle labels and pseudo labels simultaneously at each active learning iteration in an ensemble learning manner, which can be examined with IoMT Platform. Extensive experiments on multiple medical image datasets demonstrate the superiority of the proposed method over state-of-the-art active learning methods.
\end{abstract}

\begin{IEEEkeywords}
Biomedical Image Segmentation, Active Learning, Semi-supervised Learning, Ensemble Learning, Internet of Medical Things
\end{IEEEkeywords}

\section{Introduction}
\label{sec:introduction}
\IEEEPARstart{I}{mage} segmentation plays an important role in the Internet of Medical things (IoMT) systems and different applications in biology and medicine~\cite{zhao2019semi, IOMT_tao, bai2017semi}. The information from different sources including medical devices, models, doctors and sensors in the IoMT are linked by the internet or cloud services to optimize the processes for consultation, diagnosis and follow-up~\cite{scarpato2017health}. In IoMT systems, image segmentation is a vital prerequisite of computer-aided diagnosis (CAD). The past few decades have witnessed great developments in segmentation models based on deep learning and Convolutional Neural Networks (CNNs)~\cite{ronneberger2015u,2019wu,larson2017performance,SeSeNet2019} for diverse imaging modalities, such as microscopy and X-ray. However, deep learning methods heavily rely on high volumes of annotated training samples, which are difficult to obtain, especially for biomedical imaging applications. Extensive domain knowledge from biomedical experts is required for manual annotation and inspection. Besides, annotating biomedical data is costly and laborious, which results in the fact that rarely we have abundant labeled samples for biomedical image segmentation. 

In order to tackle the paucity of labeled data, some approaches different from traditional supervised learning have been developed~\cite{tajbakhsh2020embracing}. Utilizing unlabeled data, semi-supervised learning~(SSL) methods~\cite{cheplygina2019not} involve a self-training process to produce pseudo labels, in which, model updates and pseudo annotations, given the labeled data and model parameters, are performed in an alternating manner. Using unlabeled data in SSL leads to further improvement in model performance. However, there is no self-correcting process in SSL, which may cause inaccurate predictions and tend to find suboptimal solutions in the training process. Yet, a small set of accurate labels is more effective than quantities of inaccurate ones. By doing this, active learning~(AL) paradigms can be explored to select valuable samples to be labeled with high quality~\cite{dutt2016active}.

AL strategies allow enlarging the training dataset by iteratively selecting informative samples for expert annotation~\cite{bressan2018exploring}. To minimize the involvement of experts, various effective selection criteria defined on model predictions are used to discover valuable samples, which are expected to maximally boost the model performance once their annotations are obtained. AL has been extensively studied on various vision tasks~\cite{zhou2017fine}. However, there are some barriers that impede its application in clinical practice. First, the repetitive and extensive training process can harm the interactive annotation process. Second, some methods involve optimization theory and methods with high computational complexity, which further increase time costs. Third, the labeling costs are vastly different across examples, but each example is assumed to be equally expensive for the experts in most work, which leads to inefficiency and redundancy in data annotation~\cite{kuo2018cost}.

In this work, we aim to alleviate the aforementioned challenges from different perspectives. On the one hand, to reduce extra computational costs, we fully leverage the features from different hidden layers for active sample selection and avoid applying many optimization techniques. On the other hand, instead of human experts~(strong labeler) alone, we can hopefully engage machine experts~(weak labeler) to provide reliable yet cheap annotations. Therefore, we propose a novel framework that combines AL and SSL, providing synergistic annotations from both strong and weak labelers.

To reduce the complexity of the algorithm and obtain satisfactory performance, we intend to exploit the knowledge within the networks themselves, minimizing extra computational time. It is well noted that the information in hidden layers of networks can be used to supervise the learning process~\cite{lee2015deeply}. Dou~\emph{et al.}~\cite{DOU201740} first introduces the deep supervision mechanism for liver segmentation, in which, the feature maps of some hidden layers in FCN are extracted to produce dense predictions for classification error estimation. Deeply supervised learning helps speed up the training process and improve the discrimination capability. It is observed that different hidden layers of networks have different performances on the same sample, while different samples obtain different feature maps from the same layer. We hypothesize that the dissimilarity between different feature maps can be used to examine uncertainty and provide guidance for AL, so we inject deep supervision into some hidden layers of U-Net~\cite{ronneberger2015u} and design a novel AL strategy for biomedical image segmentation. The main contributions of the work are summarized as follows:

\begin{itemize}
  \item We propose a novel deep active learning framework, DSAL, which combines AL and SSL, benefiting from both strong and weak labelers. The process of expert annotation is simulated using a fully annotated dataset, while an ensemble of denseCRF~\cite{krahenbuhl2011efficient} serves as weak labelers to provide accurate pseudo labels.
  
  \item We propose a novel criterion for uncertainty evaluation, in which, multi-level segmentation results are integrated to form the basis for AL selection. In addition, a confidence criterion is designed to rectify the sample selection process and filter low-confidence samples.

  \item We evaluate DSAL on public biomedical datasets and compare it with a list of state-of-the-art AL methods. Experimental results show that DSAL significantly reduces the computational time and labeling costs without compromising the segmentation performance. 
  
\end{itemize}
\begin{figure*}[htb]
    \centering
    \includegraphics[width=\textwidth]{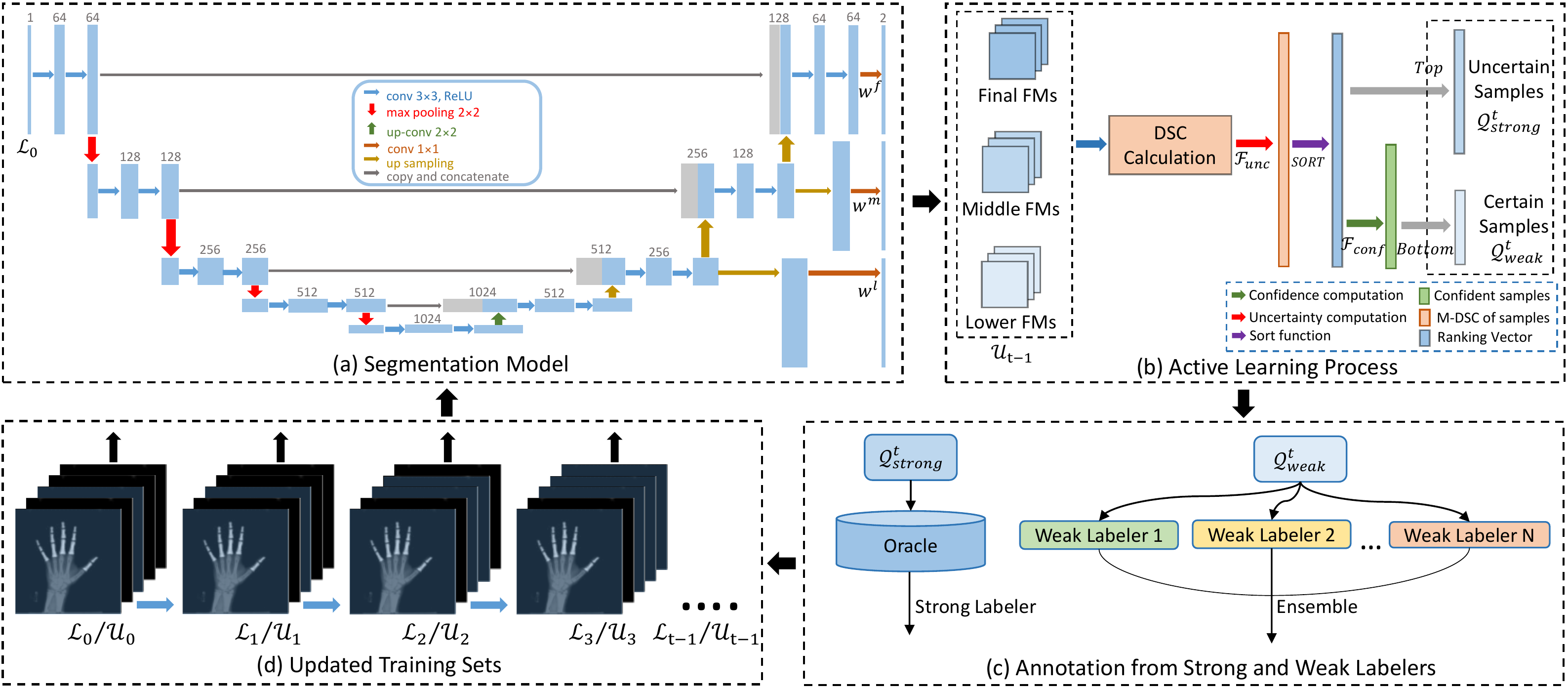}
    \caption{Overall framework of our proposed method: (a) Deeply Supervised U-Net. (b) Active Learning Process. (c) Annotation from Strong and Weak Labelers. (d) Updated Training Sets.}
    \label{fig:pipeline}
\end{figure*}

The remainder of the work is structured as follows. Section~\ref{sec:related} reviews recent work on SSL and AL. The details of DSAL are described in Section~\ref{sec:method}. The experimental datasets, setup, baselines and results are presented in Section~\ref{sec:experiments} and Section~\ref{sec:results}. Finally, Section~\ref{sec:conclusions} presents our conclusions.

\section{Related Work}~\label{sec:related}
Both semi-supervised learning and active learning aim to view the model optimization problem as a trade-off between two objectives: (i) Improve the performance and generalizability of the model among unseen data; (ii) Reduce the annotation costs in training data~\cite{tajbakhsh2020embracing}. The former focuses on better leveraging unlabeled data, while the latter seeks to wisely select informative samples to meet a given annotation budget. In this section, we give a review of the relevant literature in semi-supervised learning and active learning.

\subsection{Semi-Supervised Learning}
SSL paradigm is always considered as a self-training or pseudo-labeling process, in which, the initialized model with a small labeled dataset is further trained with pseudo annotations of unlabeled data in an iterative manner~\cite{wu2017semi}. Because no extra expert annotations are required, SSL has been popular in biomedical analysis. Bai~\textit{et al.}~\cite{bai2017semi} propose a two-step iterative method for cardiac MR image segmentation, in which, a CNN trained on the labeled data is updated with pseudo annotations refined by the application of CRF. Similarly, Zhao~\textit{et al.}~\cite{zhao2019semi} propose a student-teacher framework which consists of a U-Net~\cite{ronneberger2015u} for segmentation and a denseCRF for refinement. These methods are easy to implement but are limited by the quality of pseudo annotations. Therefore, the models can be improved by a narrow margin compared with labeled data only.

Min~\textit{et al.}~\cite{min2019two} propose a two-stream mutual attention network to reduce the noise in pseudo labels.  Xia~\textit{et al.}~\cite{xia20203d} suggest a semi-supervised framework embedded with ensemble learning. In the framework, labeled data and the corresponding ground truth are fed into multiple networks for co-training, and pseudo annotations are generated through the ensemble of network predictions. $5\%$ improvement in Dice has been achieved over a full supervised model for pancreas segmentation. These works prove that unlabeled data and pseudo labels can be helpful for performance improvement when the labeled data is limited. Moreover, combining the decisions from multiple models can help improve the quality of pseudo annotations to some extent.

\subsection{Active Learning}
AL paradigm allows extra samples to be selected and labeled. Instead of using pseudo labels, AL interactively queries new instances annotated by the oracle (expert annotator) so that the performance of networks is boosted. Therefore, the key to AL becomes how to seek informative and valuable samples for labeling. Yang~\emph{et al.}~\cite{yang_suggest} propose a two-stage framework named as Suggestive Annotation (SA), in which a set of Fully Convolutional Networks~(FCNs) with initial labeled data are trained, and samples are selected to be labeled based on (1) the uncertainty identified through an ensemble of FCNs, and (2) the similarity between high-level features extracted from images. Then, a generalized maximum coverage algorithm is proposed to suggest samples for labeling. Ozdemir~\emph{et al.}~\cite{Ozdemir_2018} also propose to select samples based on uncertainty and representativeness. Instead of multiple models, Monte Carlo Dropout (MC dropout)~\cite{gal2016dropout} is proposed to evaluate the variance of outputs from the trained network by enabling dropout at inference time. However, these frameworks consider the samples equally, regardless of the imbalanced annotation costs, which lead to higher labeling costs in practice.

Kuo~\emph{et al.}~\cite{kuo2018cost} take annotation costs into account when selecting samples, and design an AL framework based on the uncertainty of multiple PatchFCNs, in which, labeling time is estimated based on mask boundary length and number of connected components using linear regression. They formulate a knapsack $0$-$1$ problem for maximizing the uncertainty while maintaining the annotation costs. It is difficult to estimate labeling time for different modalities and tasks in biomedical imaging, for instance, segmentation on the left ventricle (LV) is more difficult than the whole heart. Another solution is to leverage pseudo labels for AL. Gorriz~\emph{et al.}~\cite{gorriz2017cost} propose a cost-effective AL method based on MC dropout, or CEAL for short. In each iteration, highly uncertain samples are selected for annotation, while samples with low uncertainties along with their predicted masks are directly used in the next iterations. CEAL is evaluated on ISIC 2017 dataset~\cite{codella2018skin} and achieved a Dice's Coefficient of $74\%$ after $9$ AL iterations.

Despite the difference in handling the labeling costs, these methods employ multiple models or variations of models to estimate the uncertainty of samples, which is slow and difficult to train, due to the repetitive training process. Besides, optimization techniques such as dynamic programming and greedy algorithm are applied in some AL methods, which further increase the computational complexity~\cite{hochbaum1996approximating, feige1998threshold}.

\section{Proposed Method}~\label{sec:method}

The proposed AL framework is illustrated in Fig.~\ref{fig:pipeline}. U-Net with deep supervision (deeply supervised U-Net, DS U-Net) is adopted as the segmentation model, where multi-level deep supervision is injected into different hidden layers. First, the model is trained on a small amount of labeled data $\mathcal{L}_{0}$, In each AL iteration $t$, the predicted masks from different layers of the same segmentation model for the remaining unlabeled data $\mathcal{U}_{t-1}$ is gathered for the AL process. In the AL process, the quality of the samples can be estimated based on the confidence and uncertainty between the results of different layers, and different samples with different qualities are dispatched to strong or weak labelers for annotation. In this way, the original training set is updated, and the segmentation model is fine-tuned with the updated dataset $\mathcal{L}_{t}$ incrementally.

\subsection{Deeply Supervised U-Net}
The segmentation model is shown in Fig.~\ref{fig:pipeline}~(a).  Different from earlier work~\cite{zhao2020deeply}, we adopt a light variant of U-Net~\cite{ronneberger2015u}, in which, upsampling layers only include nearest-neighbor interpolation instead of interpolation with $1\times1$ convolution. Upon the basic architecture of U-Net~\cite{ronneberger2015u}, we inject deep supervision into several hidden layers, namely~\textit{lower layer} and~\textit{middle layer}. Specifically, low-level and middle-level feature maps on the decoding stage are upscaled to original resolution using additional upsampling layers. Then, the softmax layer is applied to these layers to obtain predicted masks for final loss calculation. Let $W$ be the weights of the U-Net, and $w^{l}$, $w^{m}$, $w^{f}$ be the weights of three classifiers of \textit{lower layer}, \textit{middle layer} and \textit{final layer}, respectively, then the cross-entropy loss function of one layer can be formatted as:
\begin{equation}
L(\mathcal{X} ; W)=\sum_{x_{i} \in \mathcal{X}}-\log p\left(t_{i}\mid x_{i} ; W\right),
\end{equation}
where $\mathcal{X}$ denotes the training samples and $p\left(y_{i}=t\left(x_{i}\right) | x_{i} ;~ W, w^{c}\right)$ is the probability of target class label $t\left(x_{i}\right)$ corresponding to sample $x_{i} \in \mathcal{X}$, in which $c\in\{l, m, f\}$ denotes the index of the classifiers. Finally, the total loss function can be defined as:
\begin{equation}
L\left(\mathcal{X} ; W, w^{l}, w^{m}, w^{f}\right)=\sum_{c \in\{l, m, f\}} \alpha_{c} L_{c}\left(\mathcal{X} ; W, w^{c}\right), \label{eq:dsc}
\end{equation}
where $\alpha_{l}$, $\alpha_{m}$, $\alpha_{f}$ are the weights of the associated classifiers. To control the strength of different terms, in our experiments, $\alpha_{l}$, $\alpha_{m}$ and $\alpha_{f}$ are set as $0.1$, $0.3$ and $0.6$, empirically.

Finally, classifications at different levels are performed together, which helps the shallower layers to learn discriminative features more efficiently.

\subsection{Active Learning Process}
\label{sec:AL_process}

We define the AL process as follows: given a small labeled dataset $\mathcal{L}_{0}$ and unlabeled pool $U_{0}$, in each iteration $t$, find valuable samples from $U_{t-1}$ for labeling and updating the model. Therefore, the core of AL is the criterion for informative sample selection. As mentioned in previous sections, our criterion utilizes the knowledge within the networks. 

In DS U-Net, some hidden layers are supervised by optimizing the final loss function, and well-learned layer parameters always generate accurate predictions from the hidden layers. Moreover, we find that different hidden layers of the well-trained model provide consistent prediction masks for certain samples, the consistency of which can be used to evaluate the sample uncertainty. There, based on the influence of samples on the model, the predicted masks of the hidden layers (\textit{lower layer} and \textit{middle layer}) are extracted to calculate the Dice's Coefficient (DSC)~\cite{ronneberger2015u} with the results of \textit{final layer}, called \textit{L-DSC} and \textit{M-DSC}, respectively. Finally, the mean of \textit{L-DSC} and \textit{M-DSC}  serves as a proxy of the quality for each sample, termed as \textit{Mean-DSC}. Higher \textit{Mean-DSC} means a smaller uncertainty score. The uncertainty criterion is defined as $\mathcal{F}_{unc}$, which output the data with \textit{Mean-DSC}.

\begin{figure}[!htb]
    \centering
    \includegraphics[width=\linewidth]{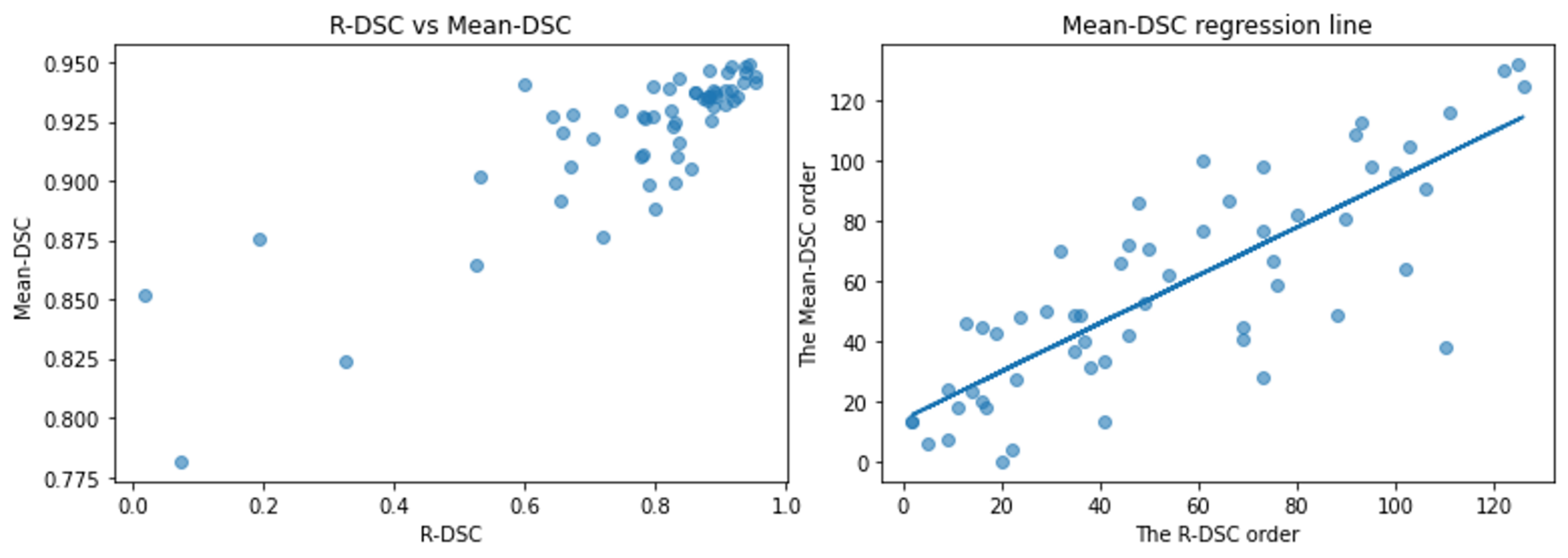}
    \caption{The relationship between R-DSC and Mean-DSC on RSNA dataset. Left: the scatter plot based on (R-DSC, Mean-DSC). Right: DSCs are sorted by descending order. Linear regression was carried out with the scatters based on (R-DSC order, Mean-DSC order). Linear regression was carried out with the scatters~(the coefficient is 0.80).}
    \label{fig:visual}
\end{figure}

Once the uncertainty score \textit{Mean-DSC} is defined, we can visualize its relation with the accuracy of the predictions. We explore the relationship between~\textit{real DSC (R-DSC)} and~\textit{Mean-DSC}, shown in Fig.~\ref{fig:visual}. In the iterative training process, we randomly select $10$ samples from each step and put them into the model for prediction, and then we sort the DSC by descending order in each iteration, and the scatter plots and trend lines between the~\textit{Mean-DSC} order and the~\textit{R-DSC} order are drawn. There is some strong relationship between these orders, which allows selecting two types of samples for labeling, as shown in the following.

\begin{enumerate}
\renewcommand{\labelenumi}{(\theenumi)}
\item \textbf{Highly uncertain samples}: Candidates with high uncertainty to be annotated by the strong labelers (the oracle).
\item \textbf{Certain samples}: The most confidently samples with high consistency, which will be annotated by the weak labelers (pseudo-labeling).
\end{enumerate}

It is noted that low-variance results can be selected based on the aforementioned uncertainty ranking strategy among different hidden layers, but the models may have high biases on the predicted masks, which means the results with low confidence may be selected for labeling. The situation becomes noticeable especially on certain samples for pseudo-labeling, as in our definition they are with low prediction uncertainty. To guarantee the high segmentation quality of the selected samples for weak labelers, we calculated the confidence of the predicted segmentation mask by averaging the pixel-wise classification confidences, which are defined as:
\begin{equation}
    \label{eq:conf}
    \mathcal{C}(x_i) = \frac{1}{H\times W} \sum_{h=0}^H \sum_{w=0}^W{\left\|\mathbf{P}_{hw} - \frac{1}{2}\right\|_1},
\end{equation}
where $\mathbf{P}\in \mathbb{R}^{H \times W}$ is the probability map of sample $x_i$. This definition presents how close the posteriors are to either background or foreground class. Therefore, the prediction confidence is effectively estimated when there is no available ground truth to be compared against. This confidence score serves as a complementary criterion to the uncertainty one which improves the selectivity of the latter.

\begin{figure}[htb]
    \centering
    \includegraphics[width=0.95\linewidth]{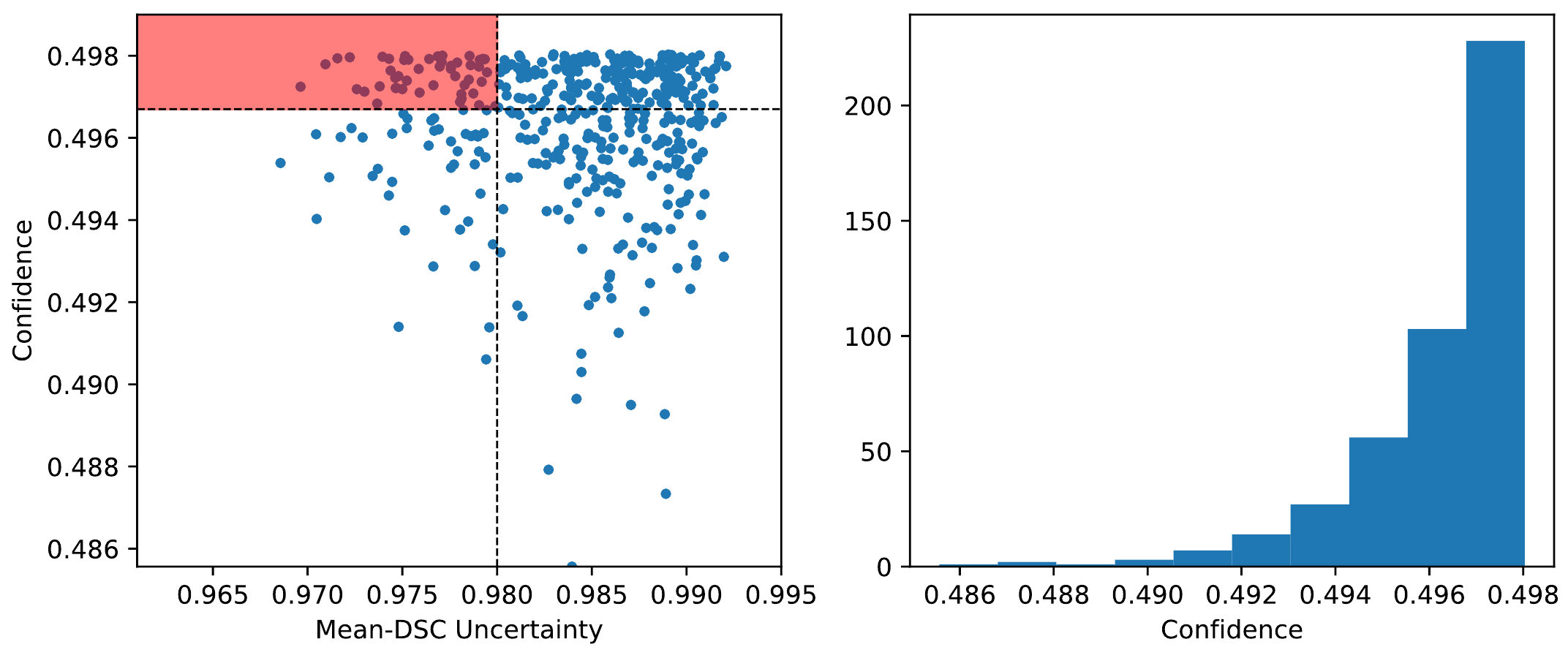}
    \caption{Illustration of the active selection process when uncertainty and confidence are involved together. (Left) Confidence vs. Mean-DSC uncertainty and (Right) histogram of the confidence scores of query samples. First samples beyond the horizontal dashed line are selected according to the confidence range of the last bin in the corresponding histogram. Then the samples are ranked based on the uncertainty function $\mathcal{F}_{unc}$ and the samples to the left of the vertical dashed line are bottom-k ones. The red area delineates the most confidently samples to be annotated by weak labelers.}
    \label{fig:unc_conf}
\end{figure}

As usual a threshold $t_{conf} \in (0, 1)$ can be fixed to filter results with low confidence for a specific model architecture and dataset. However, in the AL scenario, different AL iterations have a wide range of confidence scores. Therefore, we design a histogram-based strategy for confidence implementation. In each AL iteration, a histogram with $B$ bins can be generated by confidence scores of all remaining unlabeled samples, where only the uppermost bin is left for the following process,~\emph{i.e.}, $t_{conf}$ is adaptively set to the beginning value of the uppermost bin. The confidence criterion is formulated as:
\begin{equation}
\mathcal{F}_{conf} : \mathcal{X}\rightarrow \{ x_i\mid \mathcal{C}\left(x_{i}\right)>t_{conf}, x_{i} \in \mathcal{X} \},
\end{equation}
where $t_{\operatorname{con} f}=\frac{B-1}{B}\left(\max (\mathcal{C}(x_i)-\min (\mathcal{C}(x_i)\right)$. Fig.~\ref{fig:unc_conf} visualizes the selection process involving both uncertainty and confidence criterion for certain samples.

\subsection{Annotation from Strong and Weak Labelers}
In each AL iteration, the indices of highly uncertain samples and certain samples are exported for strong and weak labelers, respectively. We simulate the process of strong labelers using the fully annotated dataset, namely $\mathcal{F}_{\text {strong}}$. For certain samples, normally, pseudo labels generated by DS U-Net can be used in a semi-supervised learning manner. To further refine these pseudo labels, denseCRF is utilized as a weak labeler.

In feature maps, every pixel $i$ is a node with a label $x_{i}$ and an observation value $y_{i}$, while the relationships among pixels are regarded as edges. The labels $x_{i}$ behind pixels can be inferred by observations $y_{i}$, and the denseCRF $I$ is characterized by a Gibbs distribution,
\begin{equation}
P\left ( Y=y|I \right ) = \frac{1}{Z(I)}\exp{(-E(y|I))},
\end{equation}
where $E(y|I)$ is the Gibbs energy of a label $y$, which is formulated as
\begin{equation}
E(y) = \sum \limits_{i}\Psi _{u}\left ( y_{i} \right )+\sum\limits_{i<j}\Psi _{p}(y_{i},y_{j}),
\end{equation}
among which, the unary potential function $\Psi _{u}\left ( y_{i} \right )$ is donated by the output of DS U-Net, and the pairwise potentials in our model is given by
\begin{equation}
\label{eq:pairwise}
\Psi_{p}(y_{i},y_{j})=\mu(y_{i},y_{j})\sum\limits_{m=1}^{M}w^{(m)}k_{G}^{(m)}(f_{i},f_{j}),
\end{equation}
where each $k_{G}^{(m)}$ is a Gaussian kernel $k_{m}(f_{i},f_{j})$, the vectors $f_{i}$ and $f_{j}$ are feature vectors for pixels $i$ and $j$ respectively, $w^{(m)}$ are linear combination weights, and $\mu$ is a label compatibility function. In practice, it is difficult to adjust these hyperparameters in denseCRF, which may lead to large variations in the refinement process. To reduce the performance variance, we propose to implement multiple denseCRFs in an ensemble learning manner to achieve robust generalization ability. 

Specifically, we first search for a reasonably good base denseCRF specified by a set of hyperparameters including weights for both Gaussian and bilateral pairwise in Eq.~\ref{eq:pairwise}, and the times of the iteration~(step). The parameter search process is carried out using Tree-structured Parzen Estimator~\cite{bergstra2011algorithms}. Next, to introduce ensemble variance, we randomly fluctuate each of the hyperparameters around their centers (see Tab.~\ref{tab:hyperparam CRF} for central hyperparameters configurations) by sampling a sharp normal distribution around the above denseCRF, forming a set of denseCRFs slightly different from each other but with relatively good performance. We then ensemble these denseCRFs together to obtain a strong learner achieved by element-wise majority voting. In case the ensemble learner still performs suboptimal compared to the average performance of the base learners, we adopt a greedy algorithm which can fine-tune the ensemble denseCRFs by repeatedly selecting the one with the largest dice score among the base learners as a new reference for parameter fluctuation and generate new strong learners. After several rounds of such fine-tuning processes, the denseCRF initialization procedure can be ended up with the optimized set of base learners, and then their parameters are kept constant and unchanged throughout the experiments.

The pseudocode of the proposed AL process is summarized in Algorithm~\ref{algo}. In each iteration of AL, the proposed algorithm selects both uncertain samples and certain samples from unlabeled dataset $\mathcal{U}_{0}$ based on confidence score and uncertainty score. After that, the segmentation model is fine-tuned with the enlarged dataset. Then the updated model is evaluated on an out-of-bag testing dataset. The process is repeated until exhausting the cost budget or reaching satisfactory performance. We can also set a fixed iteration times $\mathcal{T}$.

\begin{algorithm}
\renewcommand{\baselinestretch}{1.5}
\newcommand\mycommfont[1]{\ttfamily\textcolor{blue}{#1}}
\SetCommentSty{mycommfont}
\SetKwInOut{KwIn}{Input}
\SetKwInOut{KwOut}{Output}
    \KwIn{Small Labeled dataset $\mathcal{L}_{0}$, Unlabeled dataset $\mathcal{U}_{0}$, iteration times $\mathcal{T}$, uncertainty function $\mathcal{F}_{unc}$, confidence function $\mathcal{F}_{conf}$, query batch size $\mathcal{K}_{strong}$, $\mathcal{K}_{weak}$, masks generation functions $\mathcal{F}_{strong}$, $\mathcal{F}_{weak}$, base model $\mathcal{M}_0$.
        }
    \KwOut{Labeled dataset $\mathcal{L}_{t}$, updated model $\mathcal{M}_{t}$.}
    \For{$t \leftarrow 1$ \KwTo $\mathcal{T}$}
    {
        \tcc{phase 1: query selection}
        $\mathcal{Q} \leftarrow \mathcal{F}_{unc}\left(\mathcal{U}_{t-1}, \mathcal{M}_{t-1}\right)$;
        
        $\mathcal{Q^\prime} \leftarrow \mathcal{F}_{conf}\left(\mathcal{Q}\right)$;

        $\mathcal{Q}_{strong}^{t} \leftarrow \{\mathcal{U}_{t-1,i}\mid i\in\mathrm{\underset{unc}{arg}}\ \mathrm{{top}K}\left(\mathcal{Q}; \mathcal{K}_{strong}\right)\}$;
        
        $\mathcal{Q}_{weak}^{t} \leftarrow \{\mathcal{U}_{t-1,i}\mid i\in\mathrm{\underset{unc}{arg}}\ \mathrm{{btm}K}\left(\mathcal{Q}^\prime; \mathcal{K}_{weak}\right)\}$;

        \tcc{phase 2: sample annotation}
        $\mathcal{\Tilde{Y}}_{strong}^{t} \leftarrow \mathcal{F}_{strong} \left(\mathcal{Q}_{strong}^{t}\right)$;
        
        $\mathcal{\Tilde{Y}}_{weak}^{t} \leftarrow \mathcal{F}_{weak} \left(\mathcal{Q}_{weak}^{t}\right)$;
        
        \tcc{phase 3: update model}
        $\mathcal{L}_{t}^{\prime} \leftarrow\mathcal{L}_{t-1} \cup\{(\mathbf{x}, y) \mid \mathbf{x} \in \mathcal{Q}_{strong}^{t}, y \in \mathcal{\Tilde{Y}}_{strong}^{t}\}$;
        
        $\mathcal{L}_{t} \leftarrow\mathcal{L}_{t}^{\prime} \cup\{(\mathbf{x}, y) \mid \mathbf{x} \in \mathcal{Q}_{weak}^{t}, y \in \mathcal{\Tilde{Y}}_{weak}^{t}\}$;

        $\mathcal{M}_{t} \leftarrow$ fine-tuning $\mathcal{M}_{t-1}$ using $\mathcal{L}_{t}$;
        
        $\mathcal{U}_{t}\leftarrow\mathcal{U}_{t-1}\backslash\left(\mathcal{Q}_{strong}^{t}\cup\mathcal{Q}_{weak}^{t}\right)$;
    }
    \KwRet{$\mathcal{L}_\mathcal{T}$, $\mathcal{M}_\mathcal{T}$.}
    \caption{The proposed active learning process}
    \label{algo}
\end{algorithm}

\section{Experiment Settings}~\label{sec:experiments}
Here, we first list and briefly describe the state-of-the-art baselines as follows:
\begin{itemize}
  \item Random Query: randomly querying $\mathcal{K}_{strong}$ samples without replacement from the unlabeled pool $\mathcal{U}_{0}$.

  \item ensemble-S~\cite{beluch2018power, yang_suggest}: an AL strategy based on uncertainty among an ensemble of $T_e$ networks. In each iteration, the mean of Shannon entropies based on multiple networks is calculated as the uncertainty score for sample selection. 

  \item ensemble-JS~\cite{kuo2018cost}: adopt similar uncertainty sampling with ensemble-S. The Jensen-Shannon divergence over the ensemble posteriors is computed.

  \item MC Dropout~\cite{gal2017deep, Ozdemir_2018}: Monte Carlo (MC) dropout is performed to obtain different class posterior probabilities in $T_m$ MC parameter sets drawn from dropout distribution, which is used to measure uncertainty.

  \item CEAL~\cite{gorriz2017cost}: a Cost-effective AL approach using dropout on unlabeled samples to estimate the pixel-wise uncertainty. Besides, unlabeled data with pseudo labels are selected for complementary sample selection.

\end{itemize}

\subsection{Datasets}

To demonstrate the effectiveness of DSAL, we carry out our experiments on two public biomedical datasets.

\begin{itemize}
  \item ISIC 2017 dataset~\cite{codella2018skin}: composed of $2000$ RGB dermoscopy images for skin lesion analysis. Each image was annotated by medical experts with pixel-wise segmentation of the melanomas. Following the AL scenario in~\cite{gorriz2017cost}, $600$ samples with annotations are randomly selected as $\mathcal{L}_{0}$, while $1000$ images are randomly selected as Unlabeled pool $\mathcal{U}_{0}$ for iterative learning. The remaining $400$ samples are used for testing.
  
  \item RSNA Bone Age dataset~\cite{larson2017performance}: includes $12611$ hand radiographs. We follow the preprocessing and sampling methods from~\cite{zhao2020deeply} and obtained a small balanced dataset where $|\mathcal{L}_{0}| = 10$ and $|\mathcal{U}_{0}| = 129$. The evaluation is carried on a striped test set of $50$ samples.

\end{itemize}

\subsection{Technical Details}

For the ISIC 2017 dataset, dermoscopic color images and corresponding binary masks are first converted into grayscale and normalized using z-score while also resized to $192\times 240$ before training. We set $\mathcal{K}_{strong} = 35$ (import $35$ oracle samples from the unlabeled pool $\mathcal{U}_{0}$ at the end of each AL iteration), while for both CEAL\cite{gorriz2017cost} and ours, pseudo-labeling process is involved from the 5th AL iteration onwards where we set both of $\mathcal{K}_{weak} = 20$ for fair comparison. We choose the dice loss as the objective function. During the training process, the base model is trained with a batch size of $32$ for $15$ epochs, and in each AL iteration, the model is fine-tuned for $10$ epochs with an enlarged dataset. The learning rate is $1\times 10^{-5}$ throughout AL training.

For the RSNA Bone Age dataset, the images are resized into $512 \times 512$, followed by standardization with min-max normalization. During the data augmentation process, the images are randomly rotated by $\pm 0.2$ degrees or flipped horizontally. We set $\mathcal{K}_{strong} = 10$ and $\mathcal{K}_{weak} = 10$ for each AL iteration. In CEAL\cite{gorriz2017cost}, the proposed algorithm samples 3 images (No-detections), 3 images (Most uncertain), and 4 images (Random) for human annotations. We choose binary cross-entropy as the objective function. In each AL iteration, the DS U-Net is trained for $150$ epochs with a batch size of $2$, and the training process will be stopped if validation loss does not improve for $30$ epochs. The initial learning rate is $1\times 10^{-4}$, which will be reduced by an order of magnitude if validation loss does not improve for $30$ epochs. We repeat the learning rate adjustment at most twice per iteration.

The experiments are conducted on an Intel Xeon Gold 6230 CPU and a Tesla V100 32GB GPU. The DS U-Net is optimized using Adam optimizer~\cite{Adam}. The selected denseCRF hyperparameters after adaptive searching are shown in Tab.~\ref{tab:hyperparam CRF}, while the number of bins in the confidence histogram for ISIC dataset and RSNA dataset are set to $10$ and $5$, respectively. For the sake of fairness, we implement the baselines by using the same segmentation architecture. In ensemble-S and ensemble-JS, an ensemble of $T_e = 3$ networks is created, while $T_m = 10$ MC parameter sets are drawn from dropout distribution.

\begin{table}[tbh]
    \centering
    \caption{Centers of hyperparameters of denseCRF for both datasets}\label{tab:hyperparam CRF}
    \begin{tabular}{c|c|c|c|c} \hline
        \multicolumn{5}{c}{ISIC} \\\hline
         & sdims & schan & compat & step \\\hline
        Gaussian Pairwise & $29.93$ & - & $9.06$ & - \\\hline
        Bilateral Pairwise & $28.19$ & $5.59$ & $9.46$ & - \\\hline
        Inference & - & - & - & $2$ \\ \hline\hline
        \multicolumn{5}{c}{RSNA} \\\hline
        Gaussian Pairwise & $1$ & - & $6$ & - \\\hline
        Bilateral Pairwise & $1$ & $7$ & $4$ & - \\\hline
        Inference & - & - & - & $1$ \\ \hline
    \end{tabular}
    
\end{table}

\section{Experimental Results}~\label{sec:results}

\subsection{Study on RSNA Bone Age dataset}

\begin{figure}[htb]
    \centering
    \includegraphics[width=0.95\linewidth]{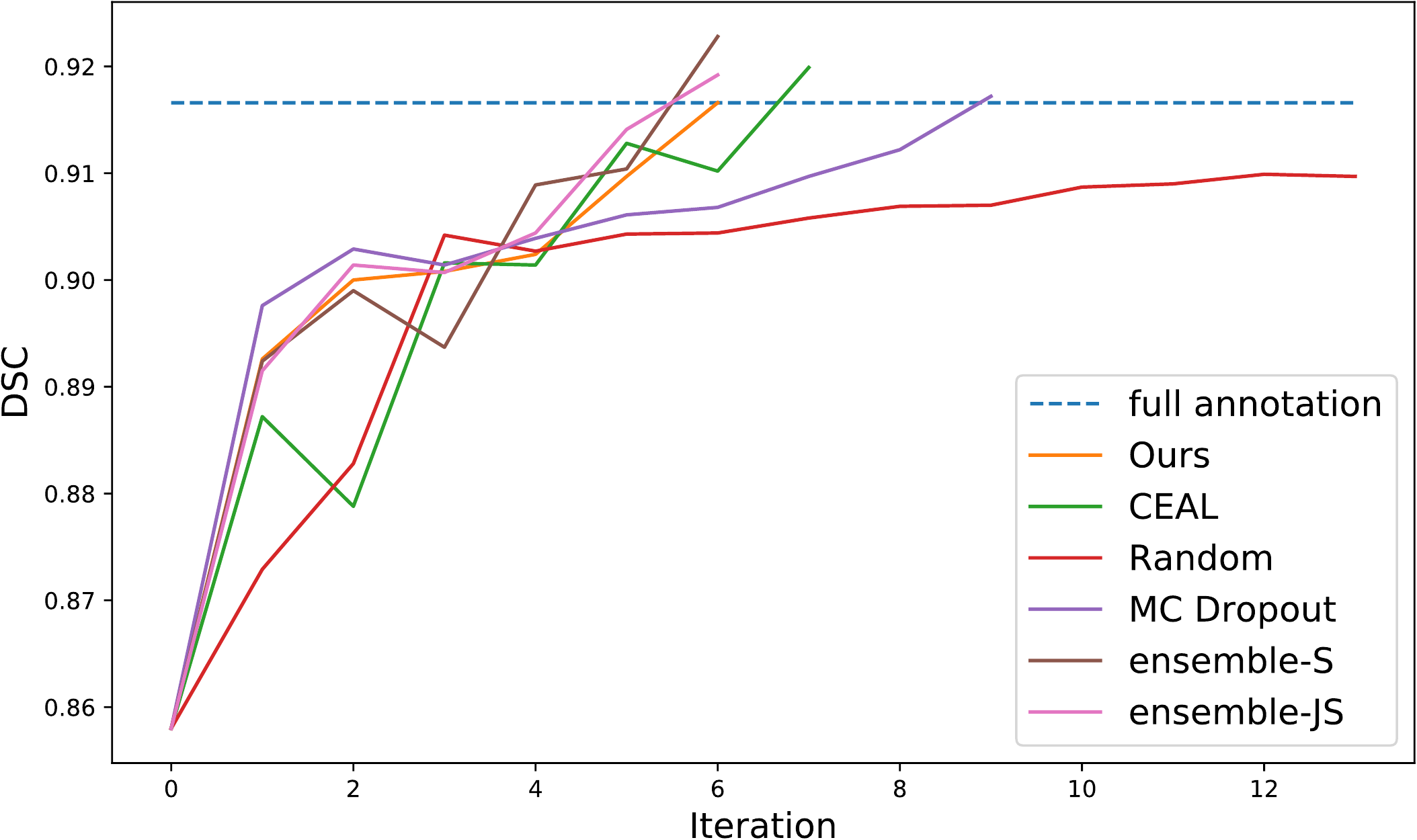}
    \caption{Experimental results on RSNA Bone Age dataset.}
    \label{fig:baselines_RSNA}
\end{figure}

The experimental results on RSNA Bone Age dataset are shown in Fig.~\ref{fig:baselines_RSNA}. We repeat the AL process for $12$ iterations until all images are annotated. We can observe that the proposed method only uses $50.36\%$ of annotations ($6$th iteration) to achieve comparable results as compared with full annotation, while other methods require more labeling resources. It is noted that the segmentation performance of ensemble-S and ensemble-JS is better than full annotation, since an ensemble of $T_e = 3$ networks always outperforms one alone. However, ensemble learning on networks is less effective and impractical for active learning, especially when $T_e$ increases dramatically. The cost of labeling is studied on ISIC 2017 dataset, and discussed in Section~\ref{result_isic}.

\subsection{Study on ISIC 2017 dataset}\label{result_isic}
\begin{figure}[htb]
    \centering
    \includegraphics[width=\linewidth]{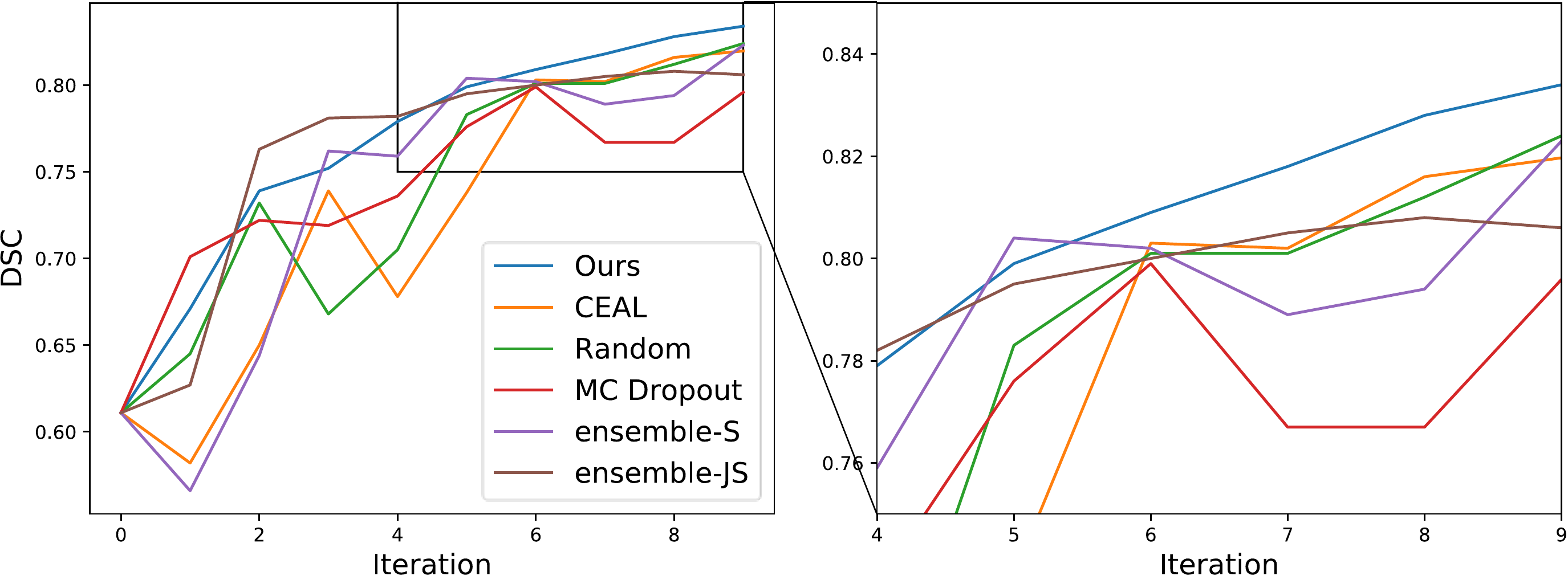}
    \caption{Experimental results on ISIC 2017 dataset (details after 4 iterations are amplified on the right side).}
    \label{fig:baselines_ISIC}
\end{figure}

The experimental results on ISIC 2017 dataset are shown in Fig.~\ref{fig:baselines_ISIC} and Tab.~\ref{tab:baselines_ISIC}. We set a fixed iteration times $\mathcal{T} = 9$ as reported in CEAL~\cite{gorriz2017cost}. The results show that the segmentation performance of the proposed method steadily increases with fewer fluctuations compared to other methods, especially in the last few iterations. It is noted that CEAL with DS U-Net reached Dice's Coefficient of 81.97 \% after 9 AL iterations, which is better than 74 \% reported in~\cite{gorriz2017cost}. This demonstrates the effectiveness of the DS U-Net. Furthermore, our method achieves the best DSC after $5$ iterations.

\begin{figure}[htb]
    \centering
    \includegraphics[width=\linewidth]{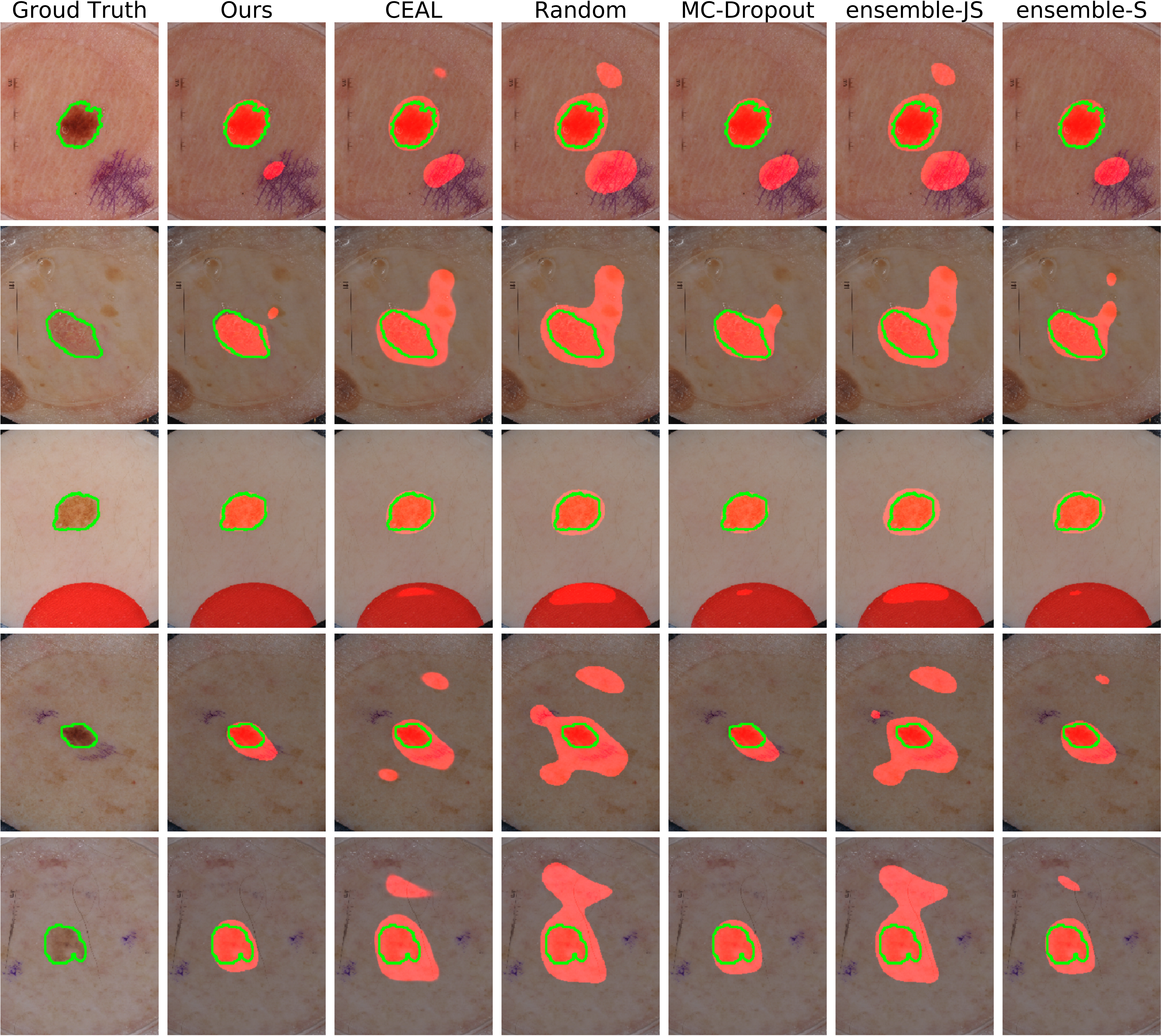}
    \caption{Segmentation of $5$ sample test images from ISIC 2017 dataset using different methods after the last AL iteration. The first column represents the original images layered with ground truth labels (bounded by green lines), and the rest show the segmentation results. The red pixels represent the segmentation masks by the corresponding methods. Ideally, pixels captured by green boundaries and red regions should be identical.}\label{fig:ISIC_vis}
\end{figure}

We visualize some segmentation results on the testing dataset of different AL models in Fig.~\ref{fig:ISIC_vis}. In consistent with the performance in Fig.~\ref{fig:baselines_ISIC}, ensemble-S can better capture the contours than ensemble-JS, yet both outperform the random query method. It is noted that the proposed method has less false positives compared to others, and generates more smooth contours. These benefit from the refinement of denseCRFs. Moreover, the involvement of multiple weak labelers can further improve the quality of pseudo labels. Besides, DS U-Net can properly grasp more meaningful and noiseless knowledge from different feature levels.
\begin{table}[htb]
    \centering
    \caption{Comparison of Dice's Coefficient on ISIC 2017 dataset among different methods across AL iterations. MCD abbreviates MC Dropout~\cite{gal2017deep,Ozdemir_2018}, ens-S abbreviates ensemble-S~\cite{beluch2018power,yang_suggest} and ens-JS abbreviates ensemble-JS~\cite{kuo2018cost}.}\label{tab:baselines_ISIC}
    
    \begin{tabular}{c||c|c|c|c|c|c} \hline
    	iter & Ours & Random & MCD & ens-S & ens-JS & CEAL~\cite{gorriz2017cost} \\\hline
        base & $0.611$ & $0.611$ & $0.611$ & $0.611$ & $0.611$ & $0.611$ \\
        $1$ & $0.671$ & $0.645$ & $\mathbf{0.701}$ & $0.566$ & $0.627$ & $0.582$ \\
        $2$ & $0.739$ & $0.732$ & $0.722$ & $0.644$ & $\mathbf{0.763}$ & $0.650$ \\
        $3$ & $0.752$ & $0.668$ & $0.719$ & $0.762$ & $\mathbf{0.781}$ & $0.739$ \\
        $4$ & $0.779$ & $0.705$ & $0.736$	& $0.759$ & $\mathbf{0.782}$ & $0.678$ \\
        $5$	& $0.799$ & $0.783$ & $0.776$ & $\mathbf{0.804}$ & $0.795$ & $0.738$ \\
        $6$ & $\mathbf{0.809}$ & $0.801$ & $0.799$ & $0.802$ & $0.800$ & $0.803$ \\
        $7$ & $\mathbf{0.818}$ & $0.801$ & $0.767$ & $0.789$	& $0.805$ & $0.802$ \\
        $8$ & $\mathbf{0.828}$ & $0.812$ & $0.767$ & $0.794$ & $0.808$ & $0.816$ \\
        $9$	& $\mathbf{0.834}$ & $0.824$ & $0.796$ & $0.823$ & $0.806$ & $0.820$ \\\hline
    \end{tabular}
\end{table}

Tab.~\ref{tab:time_ISIC} shows the average duration of all methods on the same machine. It is noted that the running time is subject to the device and its real-time workload. Given the same training strategy and networks, different active learning methods have similar training time, except ensemble-S and ensemble-JS, which require a large amount of repetitive training process, depending on $T_e$. These methods are much less practical in the real world where it could involve larger workloads on training and annotations, especially when $T_e$ is increased. We noticed that our proposed method was faster than CEAL and MC dropout methods in the AL process, while labeling time for weak labelers (denseCRFs) is 7.59 seconds per sample, which is much less than manual labeling. The time interval in our method between two iterations is acceptable, which makes it possible to deploy the method on-the-fly, especially in the setting of IoMT and edge computing.

\begin{table}[htb]
    \centering
    \caption{Average running time required in training and AL query for different methods on ISIC 2017 dataset.}\label{tab:time_ISIC}
    \begin{tabular}{c|c|c} 
    \hline
                             & training (min/sample) & AL query (min/iteration)  \\ 
    \hline
    Ours                     & $\textbf{0.079}$              & $\textbf{16.95}$                  \\ 
    \hline
    CEAL~\cite{gorriz2017cost} & $0.079$              & $88.74$                 \\
    \hline
    MCD                      & $0.079$              & $95.58 $                 \\ 
    \hline
    ens-S                    & $0.237$            & $1.05$                   \\ 
    \hline
    ens-JS                   & $0.237$              & $0.87$                   \\ 
    \hline
    Random                   & $0.079$              & $1.05$                   \\ 
    \hline
    \end{tabular}
\end{table}

\subsection{Ablation study}
To evaluate the effectiveness of each component in our holistic framework, a series of ablation experiments are carried out on ISIC dataset. As shown in Tab.~\ref{tab:ablation_ISIC}, pseudo annotations can be regarded as complementary samples in the AL scheme. When the confidence criterion is involved, the segmentation performance is improved. Finally, the performance is further boosted by the refinement of ensemble denseCRFs, which demonstrates the effectiveness of multiple weak labelers. It is clear that each component of the proposed DSAL boosts the AL performance.

\begin{table}[!htb]
\centering
\caption{Ablation study on ISIC dataset. dPL stands for selecting pseudo labels with M-DSC uncertainty measurement and ensCRF stands for ensemble denseCRF-based weak labeling. DSC is the dice score of segmentation on test samples.}\label{tab:ablation_ISIC}
\begin{tabular}{cccc||c} 
\hline
DS U-Net & dPL & Confidence & ensCRF & DSC    \\ 
\hline
$\surd$      &     &            &        & $0.8261$  \\ 
\hline
$\surd$      & $\surd$ &            &        & $0.8276$  \\ 
\hline
$\surd$      & $\surd$ & $\surd$        &        & $0.8321$  \\ 
\hline
$\surd$      &  $\surd$   &   &  $\surd$    & $0.8334$  \\ 
\hline
$\surd$      & $\surd$ & $\surd$        & $\surd$    & $0.8347$  \\
\hline
\end{tabular}
\end{table}

\section{Conclusions}~\label{sec:conclusions}

In this work, we inject SSL into the AL scheme and propose a novel AL strategy for biomedical image segmentation. In the proposed framework, two types of annotations from diverse labelers are complementary to each other, and ensemble learning is applied to improve the performance of multiple weak labelers. A novel uncertainty measurement is proposed based on the disagreement of intermediate features in the deep supervision mechanism for active sample selection. In addition, a confidence-based criterion is adopted to rectify the sample selection process for pseudo-labeling. Experimental results demonstrate that the proposed AL strategy outperforms the baselines with the inclusion of human labeling costs and computational time.

\bibliography{IEEEabrv.bib, refs}
\bibliographystyle{IEEEtran.bst}
\end{document}